%% file: iccv19_3d_dfconv_Arxiv_final.tex
\documentclass[10pt,twocolumn,letterpaper]{article}
\usepackage{iccv_teaser}
\usepackage{times}
\usepackage{epsfig}
\usepackage{graphicx}
\usepackage{amsmath}
\usepackage{amssymb}
\usepackage{color}
\usepackage{multirow}
\usepackage{booktabs}
\usepackage{array}
\usepackage{bbm}
\usepackage{epstopdf}

\usepackage[pagebackref=true,breaklinks=true,letterpaper=true,colorlinks,linkcolor=blue,citecolor=blue,bookmarks=false]{hyperref}

\iccvfinalcopy 

\def\iccvPaperID{820} %

\begin{document}

	\title{Learning Deformable Kernels for Image and Video Denoising}

\author{
	Xiangyu Xu
	\quad
	Muchen Li
	\quad
	Wenxiu Sun \\
	SenseTime Research
}

	\maketitle

	\begin{figure}[!t]\footnotesize
		\vspace{-3.3in}
		\begin{minipage}{\textwidth}
			\begin{center}
				\newcommand{\figwidth}{0.035\linewidth}
				\newcommand{\Figwidth}{0.135\linewidth}
				\newcommand{\shiftfigure}{\hspace{-3.0mm}}
				\begin{tabular}{c}
					\includegraphics[width = 0.85\linewidth]{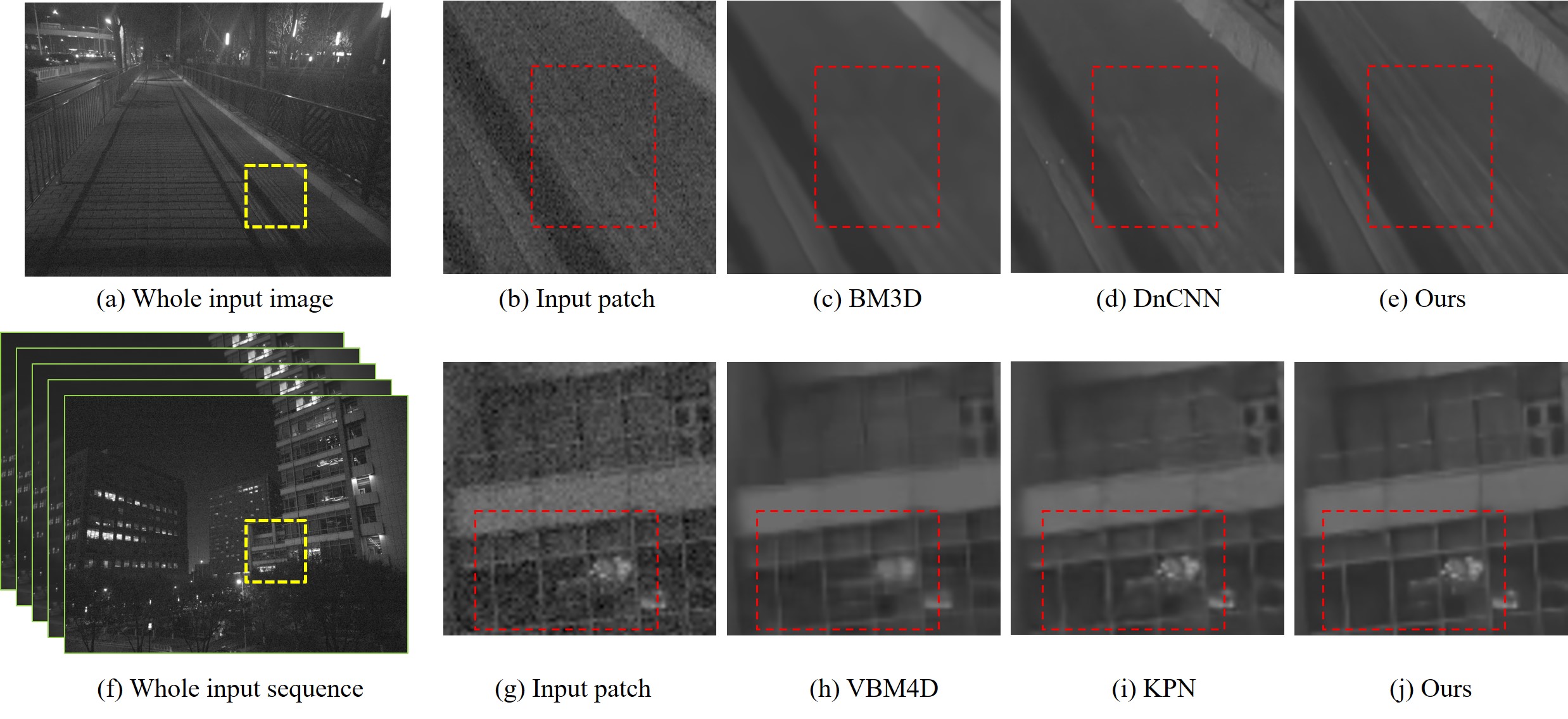} \\
				\end{tabular}
			\end{center}
			\vspace{-2mm}
			\caption{%
				Visual examples of our denoising results on image and video sequence real captured by cellphones.
				Compared to existing classical~\cite{bm3d2007,vbm4d2012} and deep learning based~\cite{dncnn,KPN} methods, 
				the proposed algorithm achieves better denoising results with less artifacts on both single image (the first row) and videos (the second row).
				(g) is cropped from the reference frame of the input sequence.}
			\label{fig:teaser} \end{minipage}
		\vspace{-6mm}
	\end{figure}

	\begin{abstract}
		Most of the classical denoising methods restore clear results by selecting and averaging pixels in the noisy input.
		Instead of relying on hand-crafted selecting and averaging strategies, we propose to explicitly learn this process with deep neural networks.
		Specifically, we propose deformable 2D kernels for image denoising where the sampling locations and kernel weights are both learned. 
		The proposed kernel naturally adapts to image structures and could effectively reduce the oversmoothing artifacts.
		Furthermore, we develop 3D deformable kernels for video denoising to more efficiently sample pixels across the spatial-temporal space.
		Our method is able to solve the misalignment issues of large motion from dynamic scenes.
		For better training our video denoising model, we introduce the trilinear sampler and a new regularization term.
		We demonstrate that the proposed method performs favorably against the state-of-the-art image and video denoising approaches on both synthetic and real-world data.
	\end{abstract}

	\vspace{-5mm}
	\section{Introduction}
	Image capturing systems are inherently degraded by noise, including shot noise of photons and read noise from sensors~\cite{healey1994radiometric}.
	And this problem gets even worse for the images and videos captured in low-light scenarios or by small-aperture cameras of cellphones. 
	Thus, it is important to study denoising algorithms to produce high-quality images and video frames~\cite{non-local-2005,bm3d2007,vbm3d2007,vbm4d2012,liu2014fast,KPN}.
	\begin{figure}[t]
		\footnotesize
		\begin{center}
			\begin{tabular}{c}
				\includegraphics[width = 0.95\linewidth]{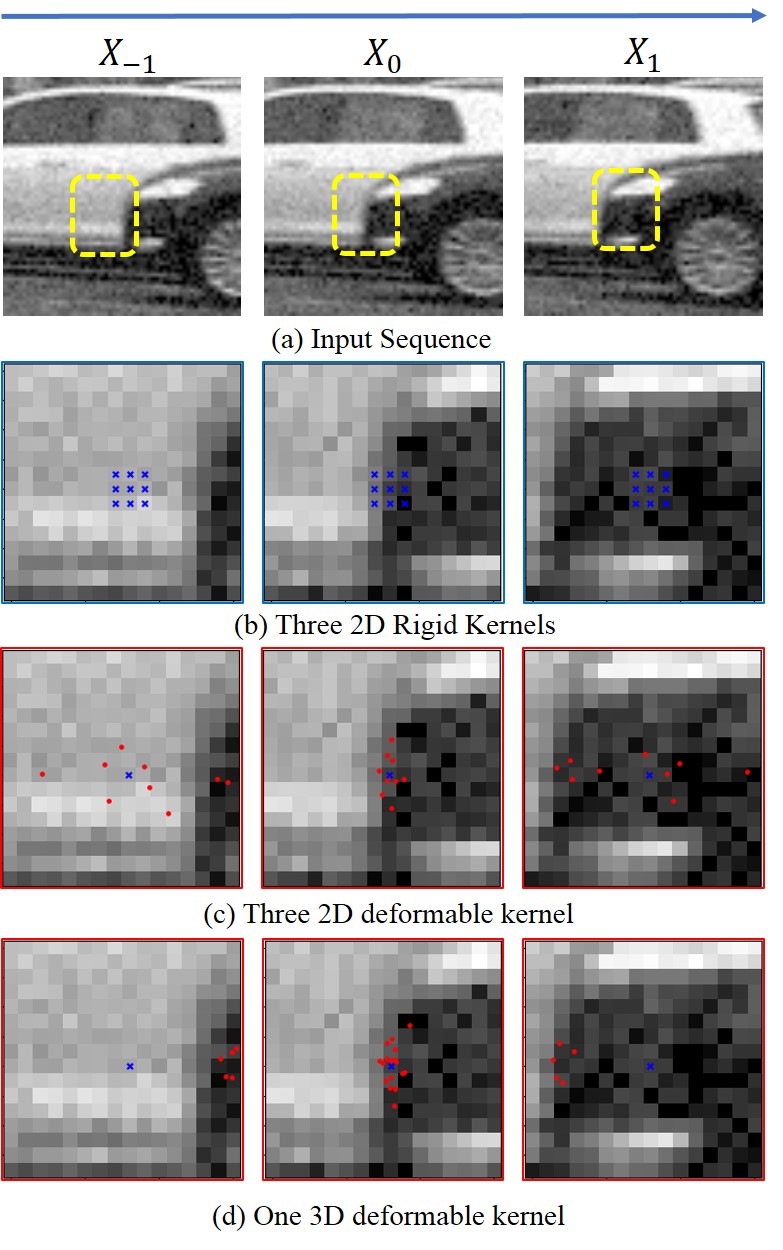} \\
			\end{tabular}
		\end{center}
		\caption{{Illustration of the denoising process using different kernels. (a) is a noisy video sequence $\{X_{-1},X_{0},X_{1}\}$.
				And the patches in the bottom are cropped from the yellow box in corresponding frames of the sequence. 
				The center blue point of patch $X_0$ in (b)-(d) indicates the reference pixel to be denoised.
				First, the rigid kernels in (b) have limited receptive field and cannot exploit the structure information in the image. 
				Moreover, it does not well handle misalignment issues in video denoising.
				Second, the proposed deformable 2D kernel (c) could adapt to image structures in $X_0$ and increase the receptive field without sampling more pixels.
				However, it does not perform well on large motion where there are few reliable pixels available in frame $X_{-1}$ and $X_{1}$.
				Third, we propose a 3D deformable kernel (d) which is able to select pixels across the spatial-temporal space, and distribute more sampling locations on more reliable frames.
			}
		}
		\label{fig:introduction}
	\end{figure}

	Most of the traditional denoising methods achieve good results by selecting and averaging pixels in the image. 
	And how to effectively select suitable pixels and compute the averaging weights are the key factors of different denoising approaches.
	As a typical example, 
	the bilateral smoothing model~\cite{tomasi1998bilateral} samples pixels in a local square region and calculates the weights with Gaussian functions.
	In addition, the BM3D~\cite{bm3d2007} searches relevant pixels by block matching, and the averaging weights are decided using empirical Wiener filter.
	However, these methods usually use hand-crafted schemes for pixel sampling and weighting, which do not always work well in complex scenarios. 

	Different from the traditional methods, deep neural networks have also been used for image denoising~\cite{dncnn,remez2017deep}.
	These data-driven models could exploit the natural image priors within large amount of training data to learn the mapping function from the noisy image to the desired clear output, which helps achieve better results than the traditional methods.
	But the deep learning based approaches do not explicitly manipulate input pixels with weighted averaging, and directly synthesizing results with the deep networks and spatially-invariant convolution kernels could lead to corrupted image textures and over-smoothing artifacts. 

	To solve the aforementioned problems, we propose to explicitly learn the selecting and averaging process for image denoising in a data-driven manner. 
	Specifically, we use deep convolutional neural networks (CNNs) to estimate a 2D deformable convolution kernel (the patch in the middle of Figure~\ref{fig:introduction}(c)) for each pixel in the noisy image. The sampling locations and weights of the kernel are both learned, corresponding to the pixel selecting and weighting strategies of the traditional denoising models, respectively.
	The advantage of our deformable kernel is two-folded.
	On one hand, the proposed approach could improve the classical averaging process by learning from data, which is by sharp contrast to the hand-crafted schemes.
	On the other hand, our model directly filters the noisy input, which constrains the output space and thereby reduces artifacts of other deep learning based approaches.
	Note that while one can simply use a normal kernel similar with the KPN~\cite{KPN} to sample pixels from a rigid grid (Figure~\ref{fig:introduction}(b)),
	it often leads to limited receptive field and cannot efficiently exploit the structure information in the images.
	And irrelevant sampling locations in the rigid kernel may harm the performance.
	By contrast, our deformable kernel naturally adapts to the image structures and is able to increase the receptive field without sampling more pixels.

	Except for the above single image case,
	we can also use the proposed method in video denoising, and a straightforward way for this is to apply the 2D deformable kernels on each frame separately, as shown in Figure~\ref{fig:introduction}(c).
	However, this simple 2D strategy has difficulties in handling videos with large motion, where few reliable pixels could be found in neighboring frames (\eg frame $X_{-1}$ and $X_1$ of Figure~\ref{fig:introduction}).
	To overcome this limitation, we need to distribute more sampling locations on the frames with higher reliability (\eg the reference frame $X_{0}$) and avoid the frames with severe motion; and this requires our algorithm to be able to search pixels across the spatial-temporal space of the input videos.
	Thus,
	instead of predicting 2D kernels for the pixels in the noisy input,
	we develop 3D deformable kernels (Figure~\ref{fig:introduction}(d)) for each pixel in the desired output to adaptively select the most informative pixels in the spatial-temporal space of videos.
	The proposed kernel naturally solves the large motion issues by capturing dependencies between 3D locations and sampling on more reliable frames, as illustrated in Figure~\ref{fig:introduction}(d).
	Furthermore,
	our method could effectively deal with the misalignment caused by dynamic scenes and reduce the cluttered boundaries and the ghosting artifacts of existing video denoising approaches~\cite{vbm4d2012,KPN} as shown in Figure~\ref{fig:teaser}.

	In this paper, we make the following contributions.
	First, we establish the connection between the traditional denoising methods and the deformable convolution, and propose a new method with deformable kernels for image denoising to explicitly learn the classical selecting and averaging process. 
	Second, we extend the proposed 2D kernels to the spatial-temporal space to better deal with large motion in video denoising, which further reduces artifacts and improves performance. We also introduce annealing terms to facilitate training the 3D kernels.
	In addition, we provide comprehensive analysis of the proposed algorithm about how the deformable kernel helps improve denoising results.
	Extensive experiments on both synthetic and real-world data demonstrate that our method compares favorably against state-of-the-arts on both single image and video inputs.
	\begin{figure*}[t]
		\footnotesize
		\begin{center}
			\begin{tabular}{c}
				\includegraphics[width = 0.9\linewidth]{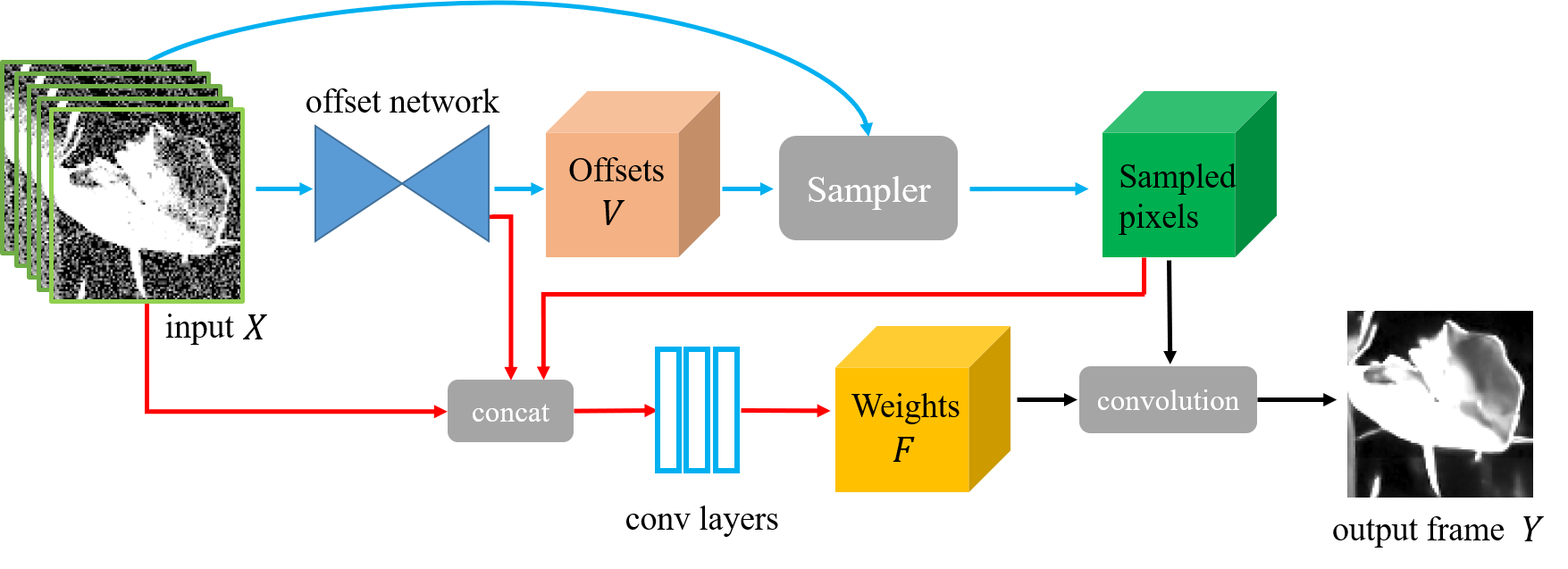} \\
			\end{tabular}
		\end{center}
		\caption{
			Overview of the proposed algorithm.
			We first learn a deep CNN (\ie the offset network) to estimate the offsets $V$ of the deformable kernels.
			And then we sample pixels from the noisy input $X$ according to the predicted offsets.
			Moreover, we concatenate the sampled pixels, the input and the features of the offset network to estimate the kernel weights $F$.
			Finally, we can generate the denoised output frame $Y$ by convolving the sampled pixels with the learned weights $F$. 
			Note that the proposed system can deal with both single image and video sequence inputs.
		}
		\label{fig:pipeline}
	\end{figure*}
	\section{Related Work}
	We discuss the state-of-the-art denoising methods as well as recent works on learning flexible convolutional operations, and put the proposed algorithm in proper context.

	\vspace{1ex}
	{\noindent \bf{Image and video denoising.}} %
	Most state-of-the-art denoising methods rely on pixel sampling and weighted averaging~\cite{gonzalez2002digital,tomasi1998bilateral,non-local-2005,bm3d2007}.
	Gaussian~\cite{gonzalez2002digital} and bilateral~\cite{tomasi1998bilateral} smoothing models use pixels from a local window and obtain averaging weights using Gaussian functions.
	NLM~\cite{non-local-2005} samples pixels globally and decides the weights with patch similarities.
	BM3D~\cite{bm3d2007} selects pixels with block matching and use transform domain collaborative filtering for averaging.
	Since video frames could provide more information then single image, the VBM3D~\cite{vbm3d2007} and VBM4D~\cite{vbm4d2012} extend the BM3D to videos by grouping more similar patches in higher dimensions.
	In addition, optical flow has been exploited in video denoising methods~\cite{liu2010high, liu2014fast}.
	However, fast and reliable flow estimation still remains a challenging problem.
	
	Deep learning has also been used for image and video denoising~\cite{dncnn,remez2017deep,rnn_denoising,godard2017deep,KPN}.
	\cite{dncnn} and \cite{remez2017deep} use deep CNN with residual connection to directly learn a mapping function from noisy input to denoised result.
	To learn the mapping function for multiple frame input,
	RNNs~\cite{rnn_denoising, godard2017deep} have also been used for exploiting the temporal structure of videos.
	While these networks are effective in removing noise, the activation functions employed in them could lead to information loss~\cite{inverting-features}, and directly synthesizing images with deep neural networks tends to cause oversmoothing artifacts.
	Mildenhall~\etal~\cite{KPN} first use deep CNN to predict normal kernels for denoising.
	However, normal kernels have rigid sampling grid and cannot well handle misalignment from camera shake and dynamic scenes.
	Instead, we propose deformable 2D and 3D kernels which enables free form pixel sampling and naturally handles above issues.
	
	\vspace{1ex}
	{\noindent \bf Learning flexible convolutions.}
	\label{sec:flexible}
	In deep CNNs, the convolutional operation is defined as weighted summation of a grid of pixels sampled from images or feature maps.
	Normal convolution kernels usually apply fixed sampling grid and convolution weights for different locations of all inputs. 
	Recently, several approaches have been developed for more flexible convolutional operations~\cite{jia2016dynamic,stn,sttn,dai2017deformable}, and the flexibility of them comes from either the weighting or the sampling scheme.
	On one hand, 
	Jia \etal~\cite{jia2016dynamic} improves the weighting strategy with a dynamic filter network.
	Following this work, similar ideas have been explored for video interpolation~\cite{niklaus2017video} and video denoising~\cite{KPN}.
	On the other hand, more flexible sampling methods have also been developed for images~\cite{stn} or videos~\cite{sttn}.
	However, these works do not allow free form convolution as each point in the desired output samples only one location.
	Dai \etal~\cite{dai2017deformable} propose spatial deformable kernels for object detection which consider 2D geometric transformations.
	But this method cannot sample pixels from the temporal space, and thereby is not suitable for video input.
	Moreover, it uses fixed convolution weights for different locations and could lead to oversmoothing artifacts like a Gaussian filter.
	By contrast, our method enables adaptively sampling in the spatial-temporal space while the kernel weights are also learned, which is consistent with the selecting and averaging process of classical denoising methods.

	\section{Proposed Algorithm}
	We propose a novel framework with deformable convolution kernels for image and video denoising.
	Different from normal kernels which have rigid sampling grid and fixed convolution weights, we use deformable grids and dynamic weights~\cite{jia2016dynamic} for the proposed kernels, which correspond to the pixel selecting and weighting process of classical denoising methods.
	The deformations of the grids could be represented as the offsets added to the rigid sampling locations.
	An overview of the proposed algorithm is shown in Figure~\ref{fig:pipeline}.
	We first train a deep CNN for estimating the offsets of the proposed kernels. 
	Then we sample pixels from the noisy input according to the predicted offsets, and estimate the kernel weights with the concatenation of the sampled pixels, the noisy input and the features of the offset network.
	Finally, we can generate the denoised output by convolving the sampled pixels with the learned kernel weights. 

		\begin{table*}[t]
		\footnotesize
		\begin{center}
			\caption{Number of feature maps for each layer of our network. We show the structure of the offset network in Figure~\ref{fig:3d_ker}(b). The ``conv layers" are presented in Figure~\ref{fig:pipeline}. ``Cn" represents the n-th convolution layer in each part of our model. $N$ is the number of sampled pixels of the proposed kernel.}
			\label{table: net_details}
			\begin{tabular}{|c|c|c|c|c|c|c|c|c|c|c|c|c|}
				\hline 
				\multirow{2}*{Layer name} & \multicolumn{10}{c|}{offset network} & \multicolumn{2}{c|} {conv layers} \\
				\cline{2-13}
				\input{results/network.tex}
				\hline
			\end{tabular}
		\end{center}
	\end{table*}

	\subsection{Learning to Sample and Average Pixels}
	\label{sec:3d kernel}
	For a noisy input $X \in \mathbb{R}^{H\times W}$ where $H$ and $W$ represent the height and width,
	the weighted averaging process~\cite{tomasi1998bilateral} for image denoising could be formulated as:
	{\small
		\begin{align}
		\label{eq:filtering}
		Y(y,x)=\sum_{n=1}^N X(y+y_n,x+x_n) F(y,x,n),
		\end{align}
	}
	where $(y,x)$ is a pixel on the denoised output $Y \in \mathbb{R}^{H\times W}$. $\{(y_n,x_n)\}$ represents the sampling grid with $N$ sampling locations, and $F\in \mathbb{R}^{H\times W \times N}$ is the weights for averaging pixels.
	For example, 
	\begin{align}
	\{(\hat{y}_n,\hat{x}_n)\}=\{(-1,-1),..,(0,0),..,(1,1)\}
	\end{align}
	defines a spatially-invariant rigid grid, where the size is $3\times3$ and $N=9$.

	In the proposed deformable kernels, the sampling grid $\{(\tilde{y}_n,\tilde{x}_n)\}$ could be generated by the predicted offsets $V \in \mathbb{R}^{H\times W \times N \times 2}$:
	{\small
		\begin{align}
		\tilde{x}_n=\hat{x}_n+V(y,x,n,1), \label{eq:V1}\\
		\tilde{y}_n=\hat{y}_n+V(y,x,n,2), \label{eq:V2}
		\end{align}
	}
	Note that both $\tilde{y}_n$ and $\tilde{x}_n$ are functions of $(y,x)$ which indicates that our deformable kernels are spatially-variant.
	Since the offsets in $V$ are usually fractional, we use bilinear interpolation to sample the pixels $X(y+\tilde{y}_n,x+\tilde{x}_n)$ similar with \cite{super-slowmo}.
	
	After the adaptive sampling, we can recover the clear output $Y$ by convolving the sampled pixels with the learned kernel $F$ as in \eqref{eq:filtering}.
	Note that the weights of $F$ are also spatially-variant and depend on the input videos, which is by contrast to the normal CNNs with fixed uniform convolution kernels.

	\begin{figure}[t]
		\footnotesize
		\begin{center}
			\begin{tabular}{c}
				\includegraphics[width = 1\linewidth]{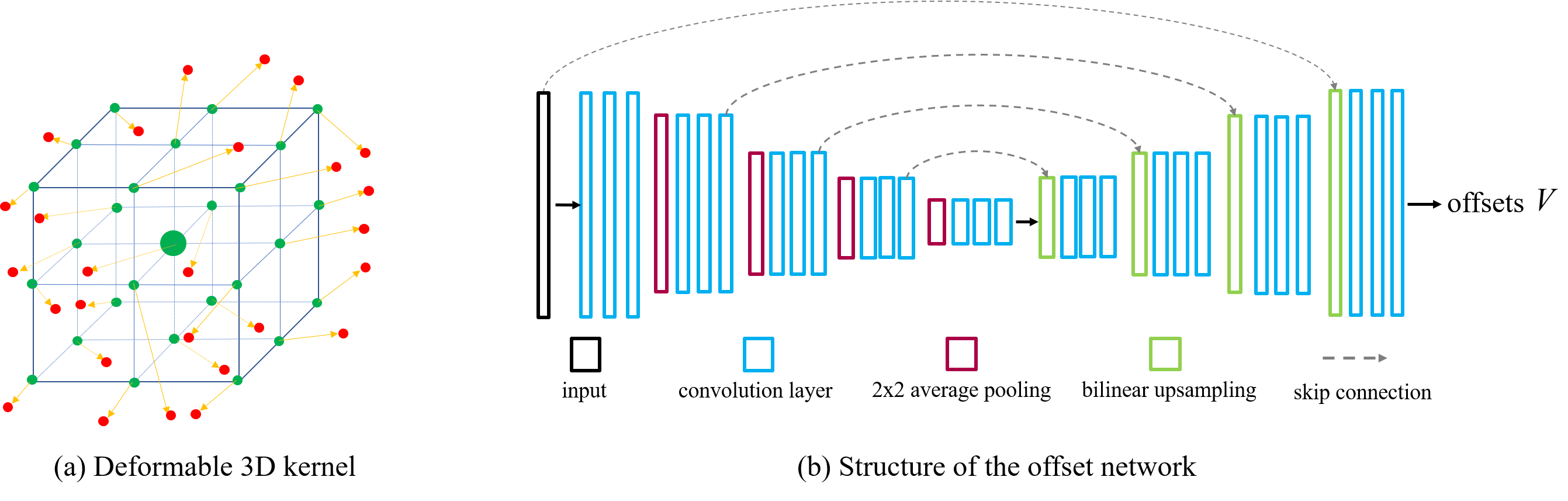} \\
			\end{tabular}
		\end{center}
		\caption{Illustration of the deformable 3D kernels (a) and the structure of the offset network (b).
		}
		\label{fig:3d_ker}
	\end{figure}

	{\flushleft\bf 3D deformable kernels.}
	We can also use the proposed method for video denoising.
	Suppose that we have a noisy video sequence : $\{X_{-\tau},...,X_0,...,X_{\tau}\}$, where $X_0$ is the reference frame.
	A straightforward way to process this input is to apply the above 2D kernels on each frame separately:
	{\small
		\begin{align}
		Y(y,x)=\sum_{t=-\tau}^\tau \sum_{n=1}^N X_t(y+y_{tn},x+x_{tn}) F_t(y,x,n).
		\end{align}
	}
	However, this simple strategy has problems in dealing with large motion as illustrated in Figure~\ref{fig:introduction}.
	To solve this problem, we develop 3D deformable kernels (Figure~\ref{fig:3d_ker}(a)) which could more efficiently distribute the sampling locations across the spatial-temporal space. 
	The 3D kernel directly takes the concatenated video frames $X \in \mathbb{R}^{H\times W \times (2\tau+1)}$ as input,
	and we can formulate the filtering process as:
	\begin{align}
	\label{eq:3d_filtering}
	Y(y,x)=\sum_{n=1}^{N'} X(y+y_n,x+x_n,t_n) F(y,x,n),
	\end{align}
	where $t_n$ denotes the sampling coordinate in the temporal dimension, and $N'$ is the number of pixels of the 3D kernel. 
	
	Similar with \eqref{eq:V1}-\eqref{eq:V2},
	we generate the sampling grid by predicting 3D offsets $V \in \mathbb{R}^{H\times W \times N' \times 3}$.
	Furthermore, to sample pixels across the video frames, we introduce the trilinear interpolation by which $X(y+y_n,x+x_n,t_n)$ could be computed as:
	{\small
		\begin{align}
		\sum_{i=1}^H \sum_{j=1}^W \sum_{t=-\tau}^\tau X(i,j,t)  \cdot max(0,1-|y+y_n-i|) \nonumber\\ 
		\cdot max(0,1-|x+x_n-j|)  \cdot max(0,1-|t_n-t|),
		\end{align}
	}
	where only the pixels closest to $(y+y_n,x+x_n,t_n)$ in the 3D space of $X$ contribute to the interpolated result.
	Since the trilinear sampling mechanism is differentiable, we could learn the deformable 3D kernels with backpropagation in an end-to-end manner.
	The derivatives of this sampler are shown in Appendix~\ref{sec:derivative}.
	
	{\flushleft\bf Gamma correction.}
	We train the denoising model in linear raw space similar with~\cite{brooks2018unprocessing}. 
	With the linear output $Y$, we further conduct gamma correction to generate the final result for better perceptual quality:
	{
		\small
		\begin{align}
		\phi(Y) & = {\begin{cases}
			12.92 Y, &Y\leq 0.0031308 \\
			(1+\alpha)Y^{1/2.4}-\alpha, & Y > 0.0031308\end{cases}}. %
		\end{align}
	}
	where  $\phi$ is the sRGB transformation function for gamma correction, and $\alpha =0.055$~\cite{anderson1996proposal}.

	\subsection{Network Architecture}
	\label{sec:network}
	The offset network in Figure~\ref{fig:pipeline} takes a single image as input for image denoising, and a sequence of $2\tau+1$ neighboring frames for video denoising.
	As shown in Figure~\ref{fig:3d_ker}(b), we adopt a U-Net architecture~\cite{ronneberger2015u} which has been widely used in pixel-wise estimation tasks~\cite{learning_to_see_in_the_dark,xu2018rendering}. 
	The U-Net is an encoder-decoder network where the encoder sequentially transforms the input frames into lower-resolution feature embeddings, and the decoder correspondingly expands the features back to full resolution estimates.
	We perform pixel-wise summation with skip connections between congruent layers in the encoder and decoder to jointly use low-level and high-level features for the estimation.
	Furthermore, we concatenate the sampled pixels, the noisy input and the features from the last layer of the offset network, and then feed them to three convolution layers to estimate the kernel weights (Figure~\ref{fig:pipeline}).
	All convolution layers use $3\times 3$ kernels with stride $1$.
	The feature map number for each layer of our network is shown in Table~\ref{table: net_details}.
	We use ReLU~\cite{nair2010rectified} as the activation function for the convolution layers except for the last one which is followed by Tanh to output normalized offsets.
	As the proposed estimation network is fully convolutional, it is able to handle arbitrary spatial size during inference.

	\subsection{Loss Function}
	With the predicted result $Y$ and ground truth image $Y_{gt}$ in linear space, we could simply use a $L_1$ loss to train our network for single image denoising:
	\begin{align}
	l(Y,Y_{gt})=\|\phi(Y)-\phi(Y_{gt})\|_1,
	\end{align}
	where gamma correction is performed to emphasize errors in darker regions and generate more perceptually pleasant results.
	{\noindent \bf{Regularization term for video denoising.}} %
	Since the deformable 3D kernel samples pixels across the video frames, 
	it is possible that the network gets stuck in the local minimum during training where all the sample locations lie around the reference frame.
	To avoid this problem and encourage the network to exploit more temporal information, 
	we introduce a regularization term to have subsets of the sampled pixels individually learn the 3D filtering. %

	Specifically, we split the $N$ sampling locations in the 3D grid $\{(y_n,x_n,t_n)\}$ into $s$ groups: $\mathcal{N}_1,...,\mathcal{N}_s$, and each group consists of $N/s$ points.
	Similar with~\eqref{eq:3d_filtering}, the filtering result of the $i$-th pixel group could be calculated as:
	\begin{align}
	Y_i(y,x) = s \sum_{j\in \mathcal{N}_i} X(y+y_j,x+x_j,t_j)F(y,x,j),
	\end{align}
	where $i\in \{1,2,...,s\}$, and the multiplier $s$ is used to match the scale of $Y$.
	With $Y_i$ for regularization, we set our final loss function for video denoising as:
	\begin{align}
	\label{eq:regu}
	l(Y,Y_{gt})+\eta \gamma^{p}l(Y_i,Y_{gt}),
	\end{align}
	The regularization of each $Y_i$ is slowly reduced during training, where the hyperparameters $\eta$ and $\gamma$ are used to control the annealing process similar with~\cite{lundy1986convergence}. 
	$p$ is the iteration number.
	At the beginning of the network optimization, $\eta \gamma^{p}\gg1$ and the second term is prominent, which encourages the network to find the most informative pixels for each subset of the 3D kernel.
	And this constraint disappears as $p$ gets larger, and the whole sampling grid learns to rearrange the sampling locations so that all the filtering groups, \ie the different parts of the learned 3D kernel, could work collaboratively.

	\begin{figure*}[t]
		\footnotesize
		\begin{center}
			\begin{tabular}{c}
				\includegraphics[width = 0.93\linewidth]{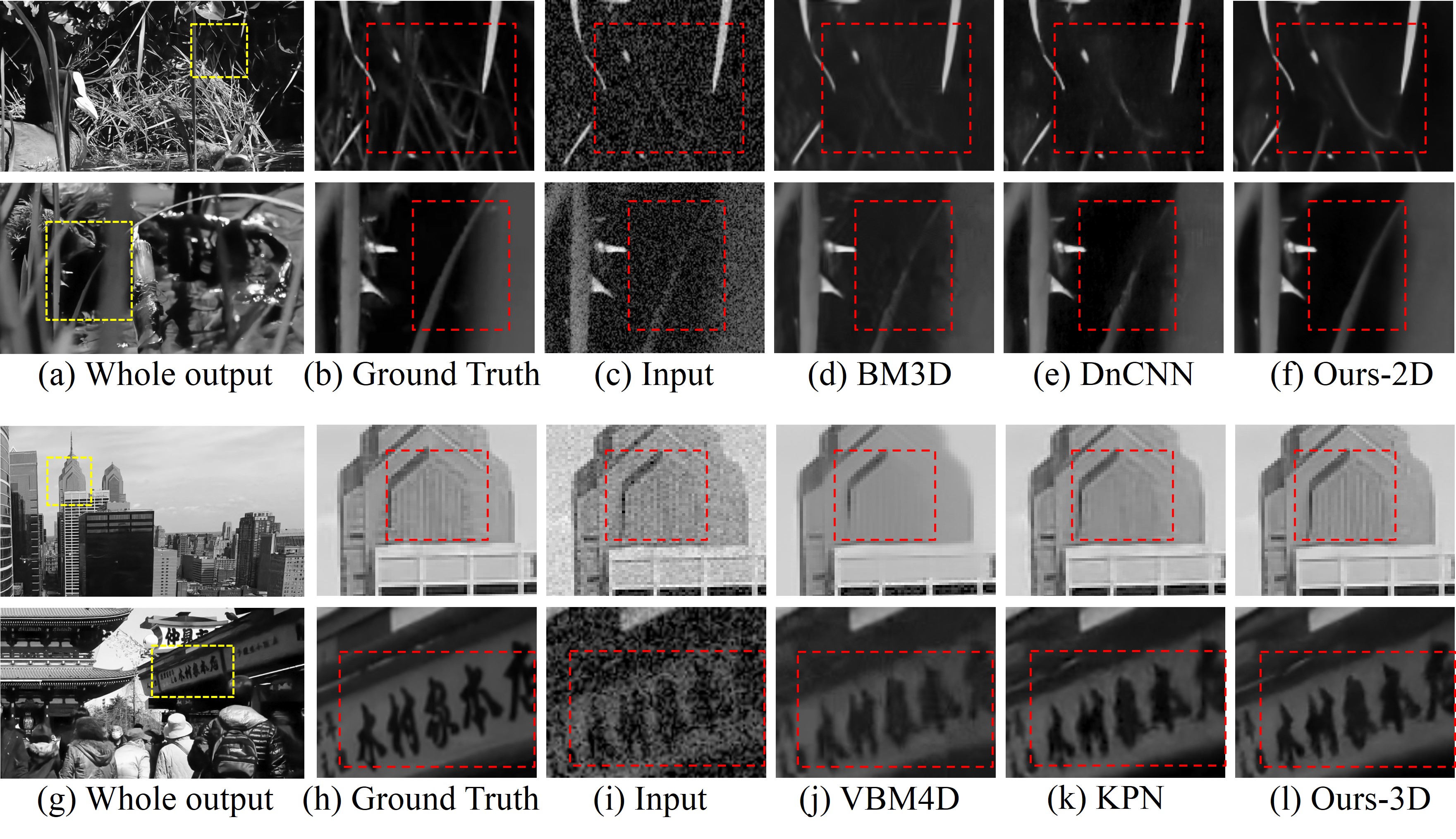} \\
			\end{tabular}
		\end{center}
		\caption{Visual examples from the synthetic dataset for single image (first and second rows) and video denoising (third and fourth rows). The proposed method achieves clearer results with less artifacts.
		}
		\label{fig:synthetic_video_1}
	\end{figure*}

	\begin{table*}[]
		\begin{center}
			\tiny
			\caption{Quantitative evaluations of single image denoising on the synthetic dataset. \#1-4 are the 4 testing subsets. ``Low" and ``High" represent different noise levels, which respectively correspond to $\sigma_s=2.5\times10^{-3},\sigma_r=10^{-2}$ and $\sigma_s=6.4\times10^{-3},\sigma_r=2\times10^{-2}$.
				{\color{red}Red} and {\color{blue}blue} indicate the first and second best performance.
				\label{table:synthetic_result_image}}
			\vspace{1mm}
			\begin{tabular}{ l |cccc|c|cccc|c }
				\hline
				\multicolumn{1}{l|}{\multirow{2}*{Algorithms}} & 
				\multicolumn{5}{c|}{Low} & 
				\multicolumn{5}{c}{High}
				\\
				\cline{2-11}
				& \#1 & \#2 & \#3 & \#4 & Average & \#1 & \#2 & \#3 & \#4 & Average \\ 
				\hline
				\input{results/results_single_image.tex}
				\hline
			\end{tabular}
		\end{center}
	\end{table*}
	
	\begin{table*}[]
		\begin{center}
			\tiny
			\caption{Quantitative evaluations of video denoising on the synthetic dataset. 
				\#1-4 are the 4 testing subsets.
				``Ours-2D" represents applying our 2D deformable kernels on each input frame separately.
				``Low" and ``High" denote different noise levels, which respectively correspond to $\sigma_s=2.5\times10^{-3},\sigma_r=10^{-2}$ and $\sigma_s=6.4\times10^{-3},\sigma_r=2\times10^{-2}$.
				{\color{red}Red} and {\color{blue}blue} text indicates the first and second best performance.
				\label{table:synthetic_result_video}}
			\vspace{1mm}
			\begin{tabular}{ l |cccc|c|cccc|c }
				\hline
				\multicolumn{1}{l|}{\multirow{2}*{Algorithms}} & 
				\multicolumn{5}{c|}{Low} & 
				\multicolumn{5}{c}{High}
				\\
				\cline{2-11}
				& \#1 & \#2 & \#3 & \#4 & Average & \#1 & \#2 & \#3 & \#4 & Average \\ 
				\hline
				\input{results/results_video.tex}
				\hline
			\end{tabular}
		\end{center}
	\end{table*}
	
	\section{Experimental Results}
	In this section, we evaluate the proposed method both quantitatively and qualitatively. 
	The source code, data, and the trained models will be made available to the public.
	{\noindent \bf{Datasets.}}
	For video denoising, we collect $27$ high-quality long videos from Internet,
	where each has a resolution of $1080\times1920$ or $720\times1280$ and a frame rate of $20$, $25$, or $30$fps.
	We use $23$ long videos for training and the other $4$ for testing,
	which are splitted into $205$ and $65$ non-overlapped scenes, respectively.
	With the videos of different scenes, we extract $20$K sequences for training where each sequence consists of $2\tau+1$ consecutive frames.
	Our test dataset is composed of $4$ subsets where each has approximately $30$ sequences sampled from the $4$ testing videos.
	The sequences used for testing are not overlapped.
	In addition, we simply use the middle frame of each sequence from the video datasets for both training and testing in single image denoising.

	Similar with~\cite{benchmarking_denoising}, we generate the noisy input for our models by first performing inverse gamma correction and then adding signal-dependent Gaussian noise: $ \mathcal{N}(0, \sigma_s q+\sigma_r^2)$,
	where $q$ represents the intensity of the pixel,
	and the noise parameters $\sigma_s$ and $\sigma_r$ are randomly sampled from  $[10^{-4},10^{-2}]$ and $[10^{-3},10^{-1.5}]$, respectively.
	In our experiments,
	we train the networks in both blind and non-blind manners. 
	For the non-blind version, the parameters $\sigma_s$ and $\sigma_r$ are assumed to be known, and the noise level is fed into the network as an additional channel of the input.
	Similar with~\cite{dncnn},
	we estimate the noise level as: $\sqrt{\sigma_r^2 + \sigma_s q_{ref} }$,
	where $q_{ref}$ represents the intensity value of the reference frame $X_0$ in video denoising or the input image in single frame denoising.
	
	Note that we only use gray scale input in our paper for fair comparisons with other methods~\cite{bm3d2007,vbm4d2012,KPN}.
	Visual examples of color images are provided in Appendix~\ref{sec: color result}.
	{\noindent \bf{Training and parameter settings.}}
	We learn $5\times5$ deformable kernels for single image denoising.
	For video input, we use $3\times3\times3$ for the deformable 3D kernels to save GPU memory.
	We set $\eta$ and $\gamma$ as $100$ and $0.9998$, respectively.
	During training, we use the Adam optimizer~\cite{kingma2014adam} with the initial learning rate as $2\times10^{-4}$.
	We decrease the learning rate by a factor of $0.999991$ per epoch, until it reaches $1\times10^{-4}$.
	We use a batch size of $32$.
	We randomly crop $128\times128$ patches from the original input for training the single image model. 
	In video denoising, we crop at the same place of all the input frames and set $\tau=2$, so that each training sample has a size of $128\times128\times5$.
	We train the denoising networks for $2\times10^5$ iterations which roughly takes 50 hours.

	{\noindent \bf{Comparisons on the synthetic dataset.}} %
	We compare the proposed algorithm with the state-of-the-art image and video denoising methods~\cite{KPN,vbm4d2012,bm3d2007,non-local-2005,dncnn} on the synthetic dataset at different noise levels. 
	We conduct exhaustive hyper-parameter finetuning for NLM~\cite{non-local-2005}, BM3D~\cite{bm3d2007} and VBM4D~\cite{vbm4d2012} including both blind and non-blind versions, and choose the best performing results.
	We also train KPN~\cite{KPN} and DnCNN~\cite{dncnn} on our datasets with the same settings.
	While \cite{KPN} is originally designed for multi-frame input, we also adapt it to single image for more comprehensive evaluation by changing the network input.

	As shown in Table~\ref{table:synthetic_result_image} and~\ref{table:synthetic_result_video}, the proposed algorithm achieves consistently better results on both single image and video denoising in terms of both PSNR and structural similarity (SSIM), compared with the state-of-art methods in all the subsets with different noise levels.
	And even our blind version model is able to achieve competitive performance, while other methods rely on the oracle noise parameters for good results. 
	In addition, KPN~\cite{KPN} uses rigid kernels for video denoising, and thus cannot effectively learn the pixel selecting process.
	And due to the inappropriate sampling locations, simply enlarging the kernel size of KPN does not lead to significant improvement as shown in Table~\ref{table:synthetic_result_image}-\ref{table:synthetic_result_video}. 

\begin{figure}[t]
		\footnotesize
		\begin{center}
			\begin{tabular}{c}
				\includegraphics[width = 0.9\linewidth]{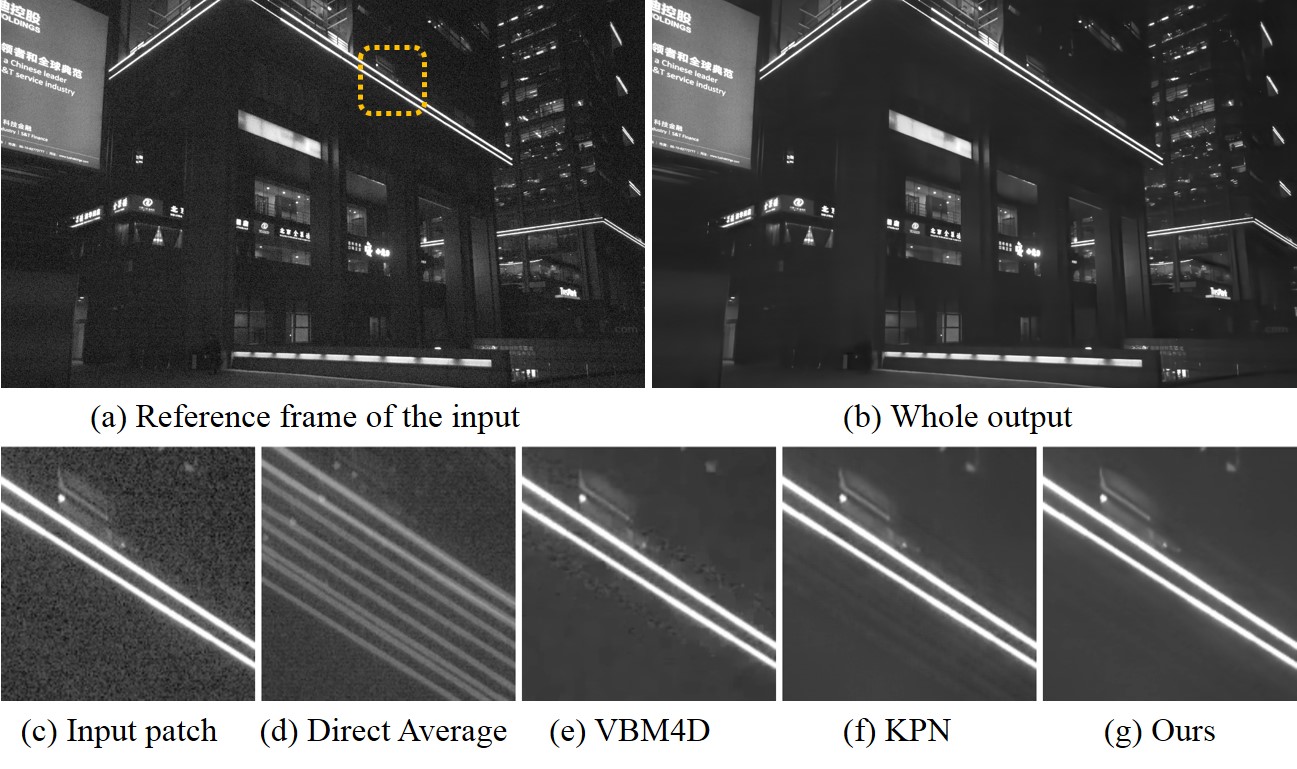} \\
			\end{tabular}
		\end{center}
		\caption{Video denoising results of a real captured sequence. (d) is generated by direct averaging the input frames. Note the ghosting artifacts around the glowing tubes of KPN in (f).
		}
		\label{fig:ghost}
	\end{figure}

	Figure ~\ref{fig:synthetic_video_1} shows examples for image and video denoising on the synthetic dataset. 
	Traditional methods~\cite{bm3d2007,non-local-2005,vbm4d2012} with hand-crafted sampling and weighting strategies do not perform well and generates severe artifacts.
	In particular, 
	VBM4D~\cite{vbm4d2012} selects pixels using $L_2$ norm to measure patch similarities, which tends to generate oversmoothing results, as shown in Figure~\ref{fig:synthetic_video_1}(j).
	On the other hand, directly synthesizing the results with deep CNNs~\cite{dncnn} could lead to corrupted structures and often lose details (Figure~\ref{fig:synthetic_video_1}(e)).
	Moreover, KPN~\cite{KPN} learns rigid kernels for video denoising, which cannot deal with misalignments larger than $2$ pixels due to the limitation of the rigid sampling.
	When the misalignment is beyond this limit, 
	the learned weights will either degenerate into a single image filter, which leads to oversmoothing result (Figure~\ref{fig:synthetic_video_1}(k)), or get trapped into disorientation, which could cause ghosting artifacts around high-contrast boundaries, as shown in Figure~\ref{fig:ghost}.
	By contrast,
	we learn the classical denoising process in a data-driven manner and achieves clearer results with less artifacts (Figure~\ref{fig:synthetic_video_1}(f), (l) and Figure~\ref{fig:ghost}(g)).
	
\begin{figure}[t]
		\footnotesize
		\begin{center}
			\begin{tabular}{c}
				\includegraphics[width = 0.9\linewidth]{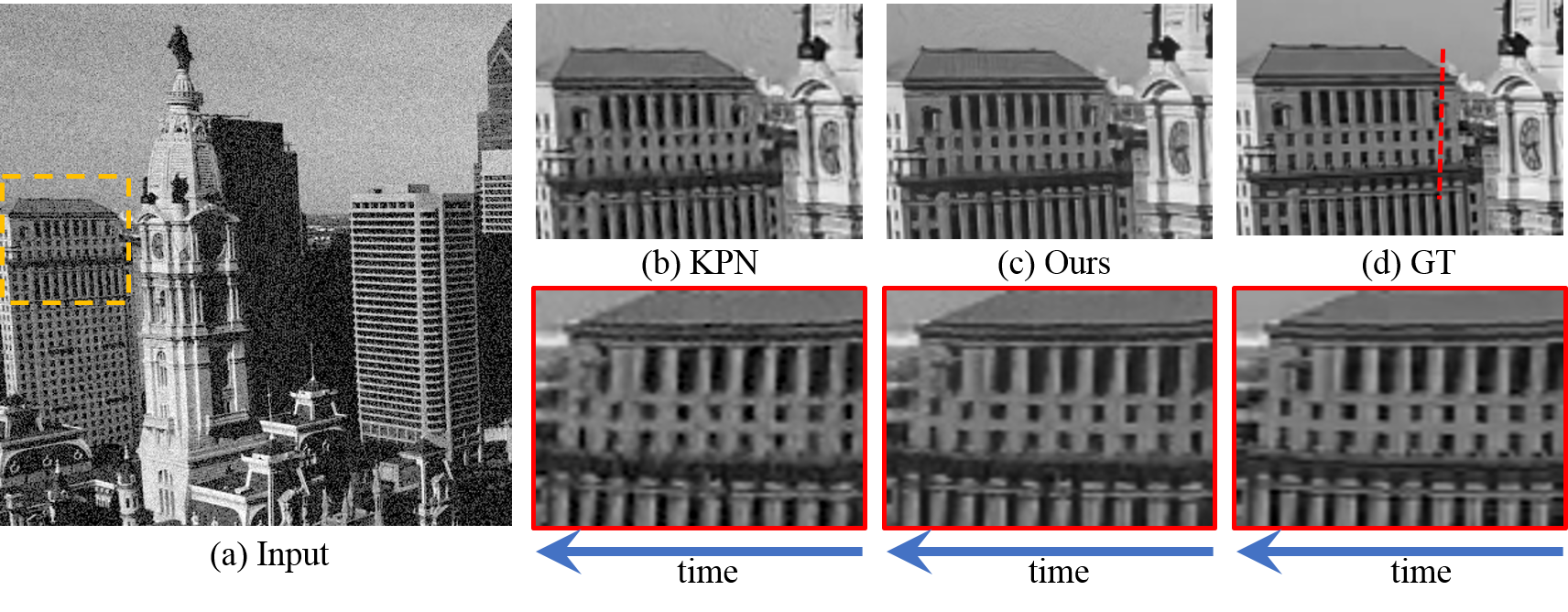} \\
			\end{tabular}
		\end{center}
		\caption{The red-box images show corresponding temporal profiles over 60 frames from the dashed line shown in the ground truth image (d). Our method achieves better temporal consistency.
		}
		\label{fig:temporal_effect}
	\end{figure}

	{\noindent \bf{Temporal consistency.}} %
	To better evaluate the results on videos,
	we show temporal comparisons in Figure~\ref{fig:temporal_effect} with the temporal profiles of the 1D sample highlighted by a dashed red line through 60 consecutive frames, which demonstrates better temporal consistency of the proposed method.

	\section{Discussion and Analysis}
	\label{sec:discuss}
	{\noindent \bf{Ablation study.}} %
	In this section, we conduct an ablation analysis on different components of our algorithm for better evaluation.
	We show the PSNR and SSIM for six variants of our model in Table~\ref{table:ablation}, where ``our full model $3\times3\times$3" is our default setting. 
	On the first row, the model "direct" uses the offset network in Figure~\ref{fig:3d_ker} to directly synthesize the denoised output  and cannot produce high-quality results.
	To learn the weighting strategies of the classical models, we use dynamic weights for the proposed deformable kernels.
	As shown in the second row of Table~\ref{table:ablation}, learning the model without dynamic weights significantly degrades the denoising performance.
	On the third and fourth rows, we learn rigid 3D kernels of different kernel sizes.
	The result shows that learning the pixel sampling strategies is important for learning the classical denoising process and improving denoising performance.
	In addition, the fifth row shows that the annealing term is also beneficial for model training, and eventually combining all components gives the best results.
	Note that the our kernel with size $3\times3\times3$ could sample pixels from a large receptive field (typically $\pm$4-$\pm$15 pixels in our experiment), and further increasing the kernel size of the deformable 3D kernel only marginally improves the performance.
	Thus, we choose smaller kernel size as our default setting to ease computational load.

	\begin{table}[]
		\begin{center}
			\footnotesize
			\caption{Ablation study on the synthetic dataset.
				\label{table:ablation}}
			\vspace{1mm}
			\begin{tabular}{ |l | cc | cc| }
				\hline
				\multicolumn{1}{|l|}{{\multirow{2}*{Algorithms}}} & 
				\multicolumn{2}{c|}{Low} & 
				\multicolumn{2}{c|}{High} \\
				\cline{2-5}
				& PSNR & SSIM & PSNR & SSIM  \\ 
				\hline
				direct &
				35.45 & 0.9518 & 32.71 & 0.9200 \\
				w/o dynamic weights & 
				35.50 & 0.9449 & 32.60 & 0.9058 \\
				fixed kernel $3\times3\times3$ &
				36.02 & 0.9555 & 33.33 & 0.9256 \\
				fixed kernel $5\times5\times5$ &
				36.37 & 0.9592 & 33.73 & 0.9320 \\
				w/o annealing term & 
				36.16 & 0.9601 & 33.48 & 0.9341 \\ 
				our full model  $3\times3\times3$ & 
				\bf 36.91 & 0.9622 & 34.23 & 0.9366 \\
				our full model  $5\times5\times5$ &
				36.88 & \bf 0.9631 & \bf 34.25 & \bf 0.9379\\
				\hline
			\end{tabular}
		\end{center}
	\end{table}

\begin{figure}[t]
		\footnotesize
		\begin{center}
			\begin{tabular}{c}
				\hspace{-4mm}
				\includegraphics[width = 0.9\linewidth]{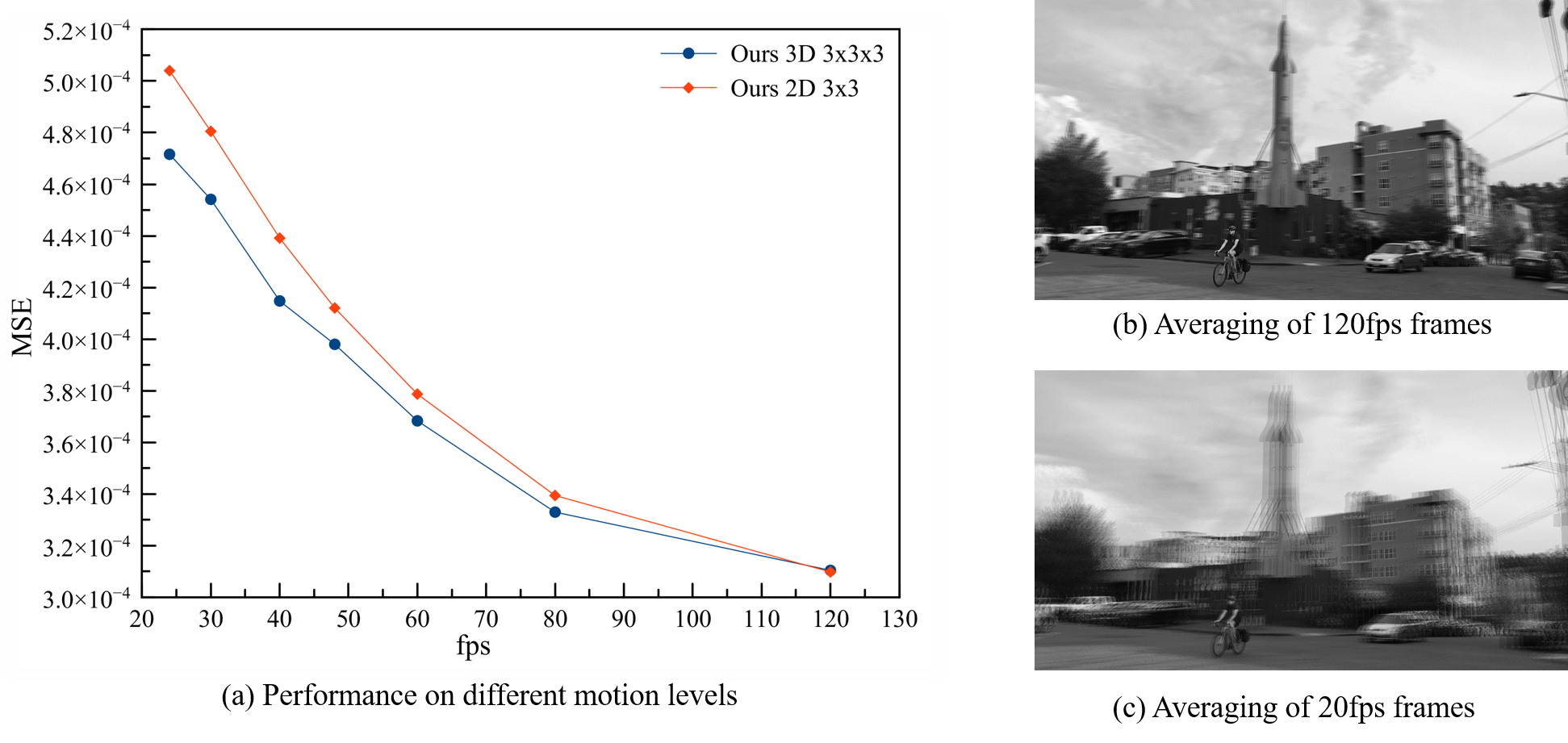} \\
			\end{tabular}
		\end{center}
		\caption{Comparison of our 2D and 3D kernels on different motion levels. Smaller frame rate at the fps-axis in (a) indicates larger motion. We visualize the motion difference by averaging the $120$fps and $24$fps input sequences in (b) and (c).
		}
		\label{fig:fps_test}
	\end{figure}

	{\noindent \bf Effectiveness of the deformable 3D kernels.}
	As illustrated in Figure~\ref{fig:introduction}, the proposed deformable 3D kernel samples pixels across the spatial-temporal space, and thus better handles large motion videos.
	To further testify the effectiveness of the 3D kernels on large motion, we compare the performance of our deformable 2D and 3D kernels under different motion levels.
	Specifically, we sample $240$fps video clips with large motion from the Adobe240 dataset~\cite{dvd}. 
	Then we temporally downsample the high frame rate videos and get $7$ test subsets of different frame rates: $120$, $80$, $60$, $48$, $40$, $30$, and $24$fps, where each contains 180 input sequences.
	Note that different frame rates correspond to different motion levels, and all the subsets use the same reference frames.
	We use one deformable 3D kernel with size $3\times3\times3$ for each pixel in the output while apply multiple $3\times3$ 2D kernels to process each frame individually.
	As shown in Figure~\ref{fig:fps_test}, the performance gap between the 2D and 3D kernels gets larger as the frame rate decreases, which demonstrates the superiority of the spatial-temporal sampling on large motion.
	We also notice that both kernels achieves better results (smaller MSE) on larger frame rates, which shows the importance of exploiting temporal information in video denoising.
	
	\begin{figure}[t]
		\footnotesize
		\begin{center}
			\begin{tabular}{c}
				\includegraphics[width = 0.9\linewidth]{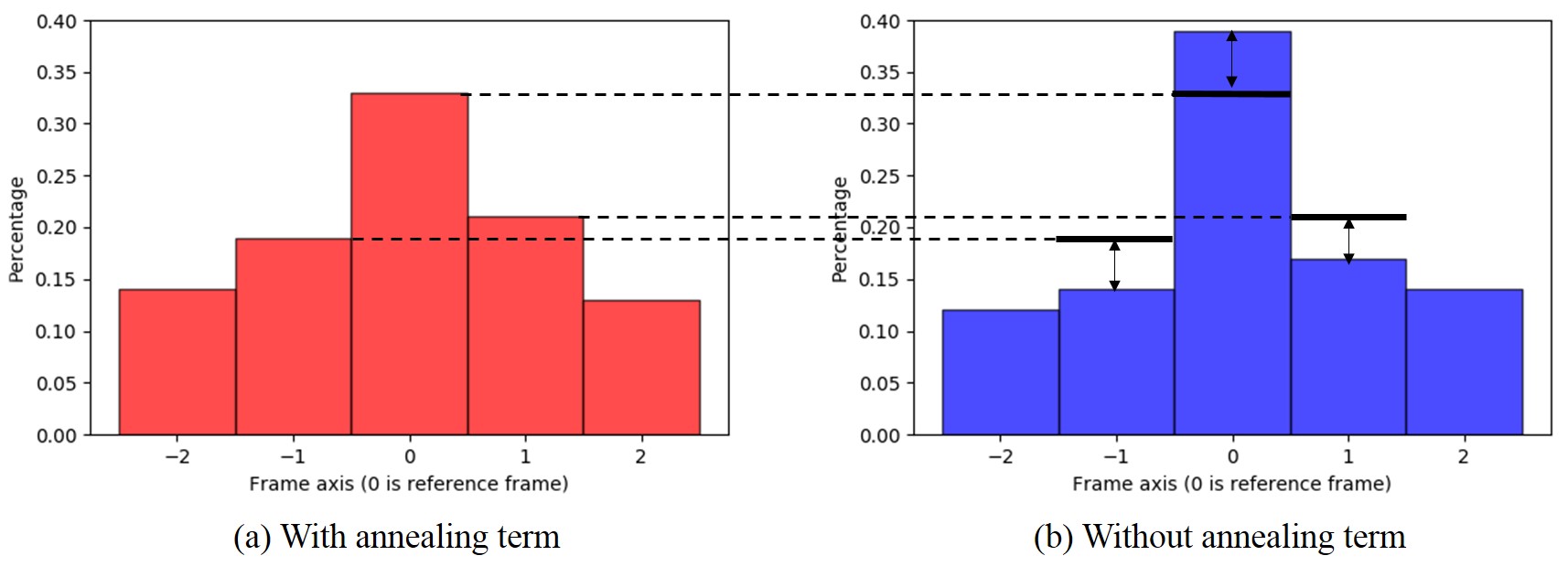} \\
			\end{tabular}
		\end{center}
		\caption{Distributions of the sampling locations on the time dimension in the test dataset. (a) and (b) represent our models with and without using the annealing term. x- and y-axis denote the frame index and the percentage of pixels, respectively.}
		\label{fig:t_distribution}
	\end{figure}

	{\noindent \bf{Effectiveness of the regularization term in \eqref{eq:regu}.}} %
	Figure~\ref{fig:t_distribution} shows the distributions of the sampling locations on the time dimension in the test dataset. 
	Directly optimizing the $L_1$ loss without the annealing term often leads to undesirable local minima where most the sampling locations are around the reference frame in video denoising as shown in Figure~\ref{fig:t_distribution}(b).
	By adding the regularization term, the network is forced to search more informative pixels across a larger temporal range, which helps avoid the local minima (Figure~\ref{fig:t_distribution}(a)).

	\begin{figure}[t]
		\footnotesize
		\begin{center}
			\begin{tabular}{c}
				\includegraphics[width = 0.95\linewidth]{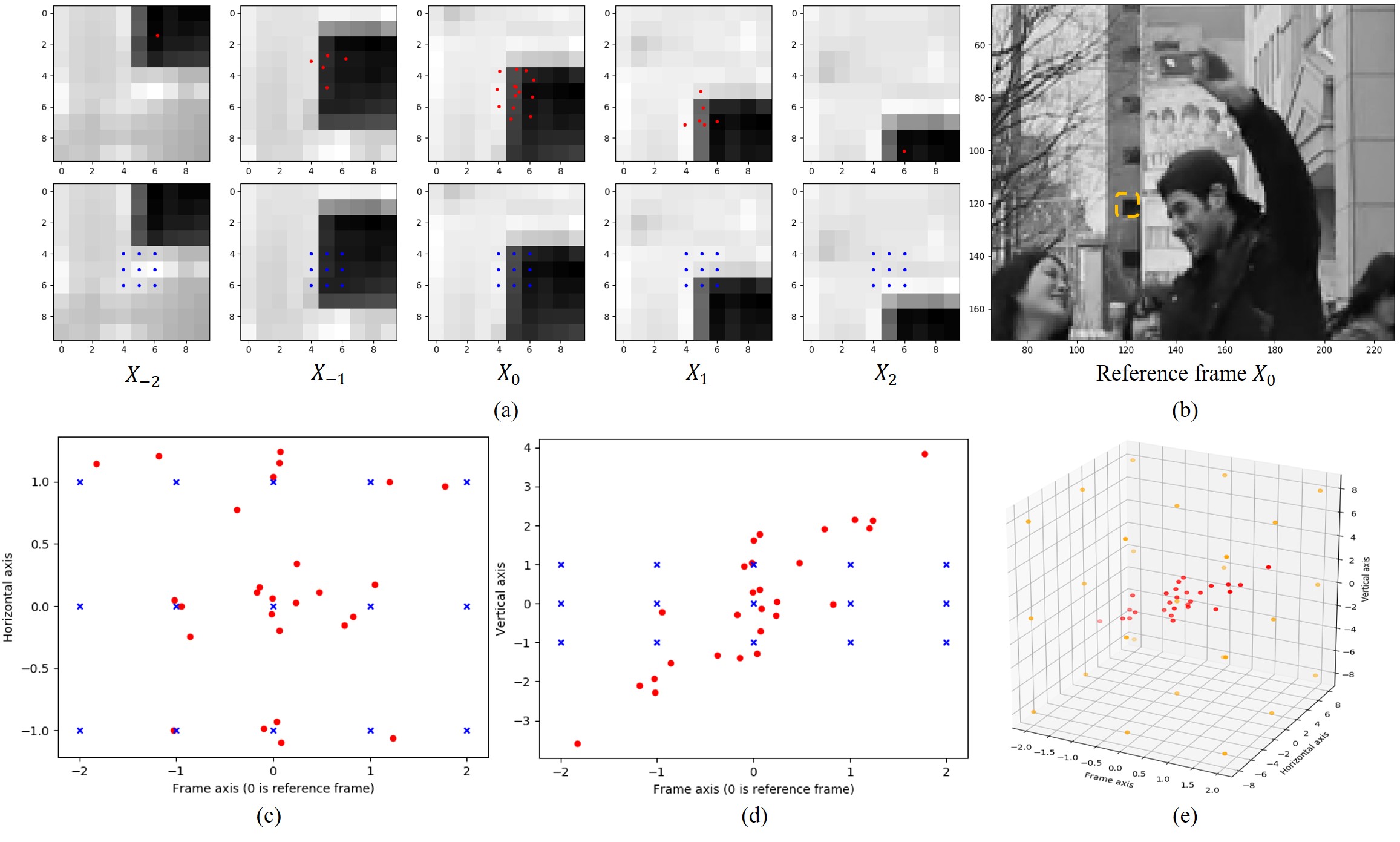} \\
			\end{tabular}
		\end{center}
		\caption{
			Visualization of the learned kernels.
			The patch sequence $\{X_{-2},X_{-1},X_0,X_{1},X_2\}$ in (a) is cropped from the same spatial location of a sequence of consecutive video frames. We show the cropping location in the original reference frame (b) with an orange curve.
			The blue points in the bottom row of (a) denote five rigid 2D kernels with size $3\times3$, while the red points in the top row of (a) represent our deformable kernel with size $3\times3\times3$.
			The center blue point in $X_0$ is the reference pixel for denoising.
			As the window in (a) is moving vertically, the sampling locations also moves vertically to trace the boundaries and search for more reliable pixels, which helps solve the misalignment issue.
			For better geometrical understanding, we show the 3D kernel in the spatial-temporal space as the red points in (e).
			In addition, we respectively project the 3D deformable kernel to different 2D planes as shown in (c) and (d).
			Note that higher coordinate in (d) indicates lower point in (a).
			With the frame index getting larger, the sampling locations distribute directionally in the Vertical axis of (d) while lying randomly in the Horizontal axis of (c), which is consistent with the motion trajectory of the window in (a).
		}
		\label{fig:learned_kernels}
	\end{figure}
\begin{figure}[t]
		\footnotesize
		\begin{center}
			\begin{tabular}{c}
				\includegraphics[width = 0.88\linewidth]{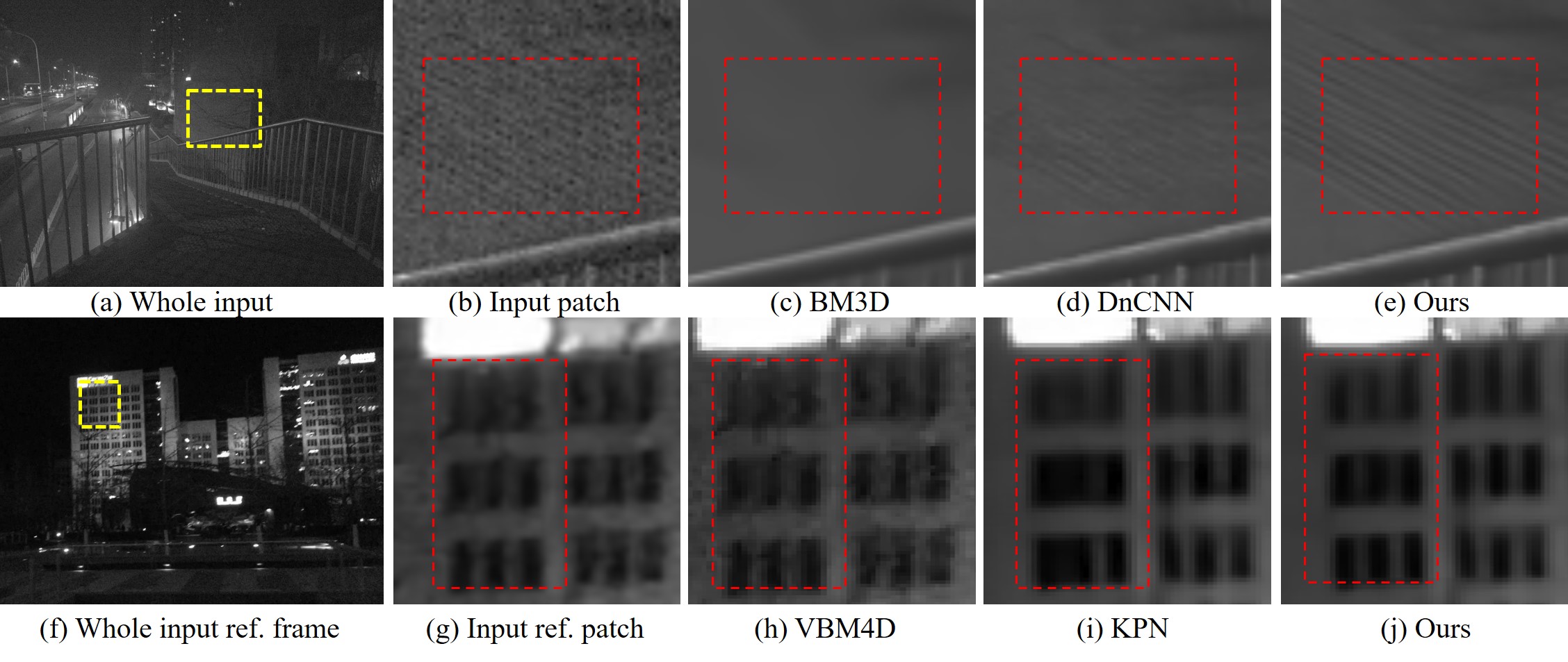} \\
			\end{tabular}
		\end{center}
		\caption{Results on real noisy image (first row) and video frame sequence (second row) captured by cellphones.
		}
		\label{fig:real_compare}
	\end{figure}

	{\noindent \bf{Visualization of the learned kernels.}} %
	We show an examples in Figure~\ref{fig:learned_kernels} to visualize the learned deformable kernels. 
	Compared to the fixed kernels (blue points in Figure~\ref{fig:learned_kernels}(a)), our deformable kernel learns  to sample pixels along the black edge across different frames, and thereby effectively enlarges the receptive field as well as reduces the interference of inappropriate pixels.
	And the ability to sample both spatially and temporally is crucial for our method to recover clean structures and details.
	{\noindent \bf{Generalization to real images.}} %
	We compare our method with state-of-the-art denoising approaches~\cite{bm3d2007,dncnn,KPN,vbm4d2012} on real images and video sequences in Figure~\ref{fig:teaser} and \ref{fig:real_compare} captured by cellphones.
	While trained on synthetic data, 
	our model is able to recover subtle edges from the real-captured noisy input and well handles misalignment from large motions.

	\section{Conclusions}
	In this work, we propose to explicitly learn the classical selecting and averaging process for image and video denoising with deep neural networks.
	The proposed method is able to adaptively select pixels from the 2D or 3D input, which well handles misalignment from dynamic scenes and enables large receptive field while preserving details.
	In addition, we introduce new techniques for better training the proposed model.
	We can achieve a running speed of 0.27 megapixels per second on a GTX 1060 GPU, while the speed is 0.3 for KPN~\cite{KPN}. 
	We will explore more efficient network structures for estimating the denoising kernels in future work.

	\appendix
	
	\section{Derivatives of the Deformable Convolution}
	\label{sec:derivative}
	As introduced in Figure~\ref{fig:pipeline}, the result $Y$ of the deformable convolution depends on the learned kernel weights $F$ and sampling offsets $V$. 
	For training the proposed neural network, we need the derivatives with respect to both $F$ and $V$ which could be derived as follows:
	
	\begin{align}
	\frac{\partial Y(y,x)}{\partial F(y,x,n)} = & X(y+y_n,x+x_n,t_n)\\
	\frac{\partial Y(y,x)}{\partial V(y,x,n,1)} = & F(y,x,n) \cdot \sum_{i=1}^H \sum_{j=1}^W \sum_{t=-\tau}^\tau X(i,j,t) \nonumber\\ 
	&\cdot max(0,1-|y+y_n-i|) \nonumber\\
	&\cdot max(0,1-|t_n-t|)   \nonumber\\
	&\cdot \left\{
	\begin{array}{lll}       %
	0, \text{~if~} |x+x_n-j| \geq 1 \\  %
	1, \text{~if~} -1<x+x_n-j<0 \\  %
	-1, \text{~otherwise}
	\end{array}              %
	\right. \\
	\frac{\partial Y(y,x)}{\partial V(y,x,n,2)} = & F(y,x,n) \cdot \sum_{i=1}^H \sum_{j=1}^W \sum_{t=-\tau}^\tau X(i,j,t) \nonumber\\ 
	&\cdot max(0,1-|x+x_n-j|) \nonumber\\ 
	&\cdot max(0,1-|t_n-t|)   \nonumber\\ 
	&\cdot \left\{
	\begin{array}{lll}       %
	0, \text{~if~} |y+y_n-i| \geq 1 \\  %
	1, \text{~if~} -1<y+y_n-i<0 \\  %
	-1, \text{~otherwise}
	\end{array}              %
	\right. \\
	\frac{\partial Y(y,x)}{\partial V(y,x,n,3)} = & F(y,x,n) \cdot \sum_{i=1}^H \sum_{j=1}^W \sum_{t=-\tau}^\tau X(i,j,t) \nonumber\\ 
	&\cdot max(0,1-|y+y_n-i|) \nonumber\\ 
	&\cdot max(0,1-|x+x_n-j|) \nonumber\\ 
	&\cdot \left\{
	\begin{array}{lll}       %
	0, \text{~if~} |t_n-t| \geq 1 \\  %
	1, \text{~if~} -1<t_n-t<0 \\  %
	-1, \text{~otherwise}
	\end{array}              %
	\right. 
	\end{align}
	
	\section{Results on Color Videos}
	\label{sec: color result}
\begin{figure}[h]
		\footnotesize
		\begin{center}
			\includegraphics[width = 1\linewidth]{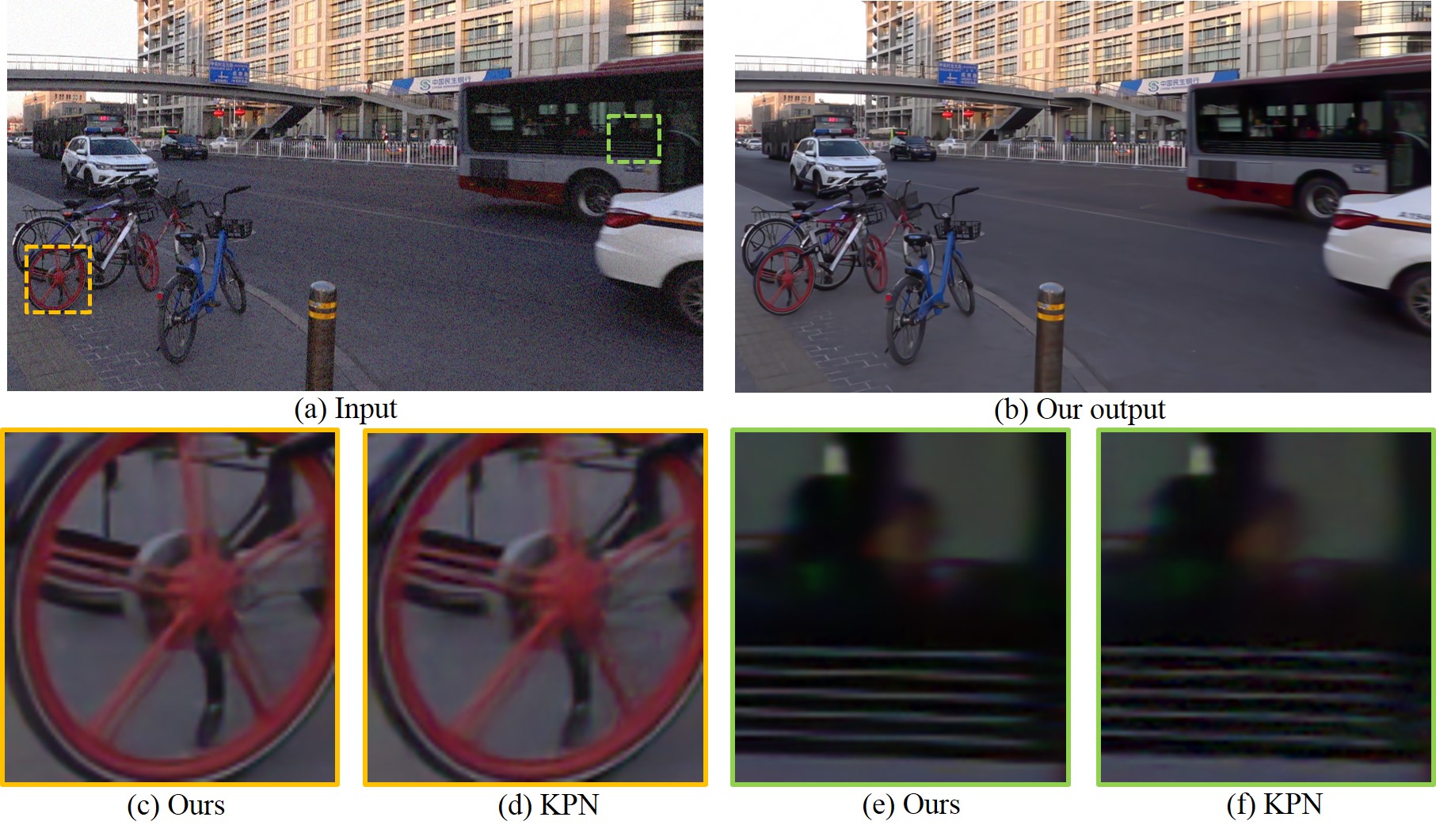} 
		\end{center}
		\caption{Qualitative evaluation on color videos. Our method generates clearer frames with less artifacts compared with KPN.}
		\label{fig:color}
	\end{figure}

	Since KPN~\cite{KPN} only considers grayscale videos, we also only use gray input in our paper for fair comparisons.
	To further evaluate the proposed deformable kernel on color videos, we respectively process the R, G, B channels with our network and provide a color example in Figure~\ref{fig:color} for comparison. Note that our result has less color artifacts around edges.
	
	{\small
		\bibliographystyle{ieee}
		\bibliography{egbib}
	}

\end{document}

%% file: results/network.tex
       & C1-3 & C4-6 & C7-9 & C10-12 & C13-15 & C16-18 & C19-21 & C22-24 & C25-26 & C27 & C1-2 & C3 \\
\hline
number of feature maps & 64 & 128 & 256 & 512 & 512 & 512 & 256 & 128 & 128 & $N\times$3 & 64 & $N$ \\

%% file: results/results_single_image.tex
Reference frame 
&26.75 / 0.6891&28.08 / 0.7333&27.37 / 0.5842&27.96 / 0.7064& 27.54 / 0.6782
&22.83 / 0.5403&23.94 / 0.5730&23.00 / 0.3746&23.97 / 0.5598& 23.43 / 0.5119\\
NLM~\cite{non-local-2005} 
& 31.04 / 0.8838 & 31.51 / 0.9025 & 33.35 / 0.8687 & 31.71 / 0.8663 & 31.90 / 0.8803 
& 28.21 / 0.8236 & 28.57 / 0.8443 & 30.62 / 0.8076 & 28.73 / 0.8040 & 29.03 / 0.8199\\
BM3D~\cite{bm3d2007} 
& 33.00 / 0.9196 & 32.63 / 0.9245 & 35.16 / 0.9172 & 33.09 / 0.9028& 33.47 / 0.9160
& 29.96 / 0.8793 & 29.81 / 0.8836 & 32.30 / 0.8766 & 30.27 / 0.8609& 30.59 / 0.8751 \\
DnCNN~\cite{dncnn} 
& 35.30 / 0.9499&	34.54 / 0.9498&	37.45 / 0.9436&	36.22 / 0.9494&	35.88 / 0.9482&	32.30 / 0.9163&	31.54 / 0.9124&	34.55 / 0.9048&	33.26 / 0.9148&	32.91 / 0.9121 \\
KPN~\cite{KPN}, $5\times5$ 
&35.23 / 0.9526& 34.38 / 0.9493& 37.50 / 0.9451&	36.18 / 0.9526&	35.82 / 0.9499& 32.32 / 0.9198& 31.44 / 0.9120&	34.74 / 0.9085&	33.28 / 0.9200& 32.94 / 0.9151 \\
KPN~\cite{KPN}, $7\times7$ 
& 35.23 / \color{blue}0.9534& 34.38 / 0.9500& 37.54 / 0.9460&	36.16 / 0.9536&	35.83 / 0.9508&	32.36 / 0.9222&	31.46 / 0.9136&	34.80 / \color{blue}0.9110&	33.30 / \color{blue}0.9220&	32.98 / 0.9172 \\
Ours-2D 
& {\color{red}35.40} / \color{red}0.9535& {\color{red}34.57} / \color{blue}0.9507& {\color{red}37.64} / \color{red}0.9465&	{\color{red}36.41} / \color{red}0.9538&	{\color{red}36.01} / \color{red}0.9511
& {\color{red}32.49} / \color{red}0.9226& {\color{red}31.62} / \color{red}0.9153& {\color{red}34.89} / \color{red}0.9121& {\color{red}33.51} / \color{red}0.9232& {\color{red}33.13} / \color{red}0.9183 \\
\hline
KPN~\cite{KPN}, $5\times5$, $\sigma$ blind 
& 35.18 / 0.9492&	34.20 / 0.9484&	37.39 / 0.9438&	36.05 / 0.9508&	35.71 / 0.9480&	32.23 / 0.9182&	31.37 / 0.9107&	34.63 / 0.9073&	33.17 / 0.9183&	32.85 / 0.9136\\
DnCNN, $\sigma$ blind 
& 35.19 / 0.9500&	34.38 / 0.9479&	37.28 / 0.9417&	36.06 / 0.9491&	35.73 / 0.9472&	32.19 / 0.9158&	31.42 / 0.9105&	34.40 / 0.9023&	33.08 / 0.9135&	32.77 / 0.9105\\
Ours-2D, $\sigma$ blind 
& {\color{blue}35.33} / 0.9531& {\color{blue}34.55} / \color{red}0.9508& {\color{blue}37.57} / \color{blue}0.9458& {\color{blue}36.35} / \color{red}0.9538& {\color{blue}35.95} / \color{blue}0.9509
& {\color{blue}32.44} / \color{blue}0.9224&	{\color{red}31.62} / \color{blue}0.9152& {\color{blue}34.81} / 0.9109& {\color{blue}33.46} / 0.9215& {\color{blue}33.08} / \color{blue}0.9175\\

%% file: results/results_video.tex
Direct average 
&22.75 / 0.6880& 25.70 / 0.7777& 25.15 / 0.6701& 23.47 / 0.6842& 25.27 / 0.7050
&21.96 / 0.6071& 24.78 / 0.6934& 24.34 / 0.5466& 22.81 / 0.6055& 23.47 / 0.6132\\
VBM4D 
&33.26 / 0.9326 & 34.00 / 0.9469& 35.83 / 0.9347& 34.01 / 0.9327 &34.27 / 0.9367
&30.34 / 0.8894 & 31.28 / 0.9089&32.66 / 0.8881& 31.33 / 0.8925
&31.40 / 0.8947\\
KPN~\cite{KPN}, $5\times5$
&35.61 / 0.9597&35.25 / 0.9637&38.18 / 0.9529&36.45 / 0.9604&36.37 / 0.9592
&32.92 / 0.9344&32.56 / 0.9358&35.59 / 0.9223&33.80 / 0.9355&33.72 / 0.9320\\
KPN~\cite{KPN}, $7\times7$
&35.64 / \color{blue}0.9603&35.23 / 0.9646&38.30 / \color{blue}0.9542&36.49 / \color{blue}0.9623&36.41 / \color{blue}0.9604
&32.95 / \color{blue}0.9336&32.61 / 0.9377&35.72 / \color{blue}0.9246&33.88 / \color{blue}0.9374&33.79 / \color{blue}0.9333\\
Ours-2D, $3\times3$
&35.66 / 0.9576&{\color{red}35.82} / \color{blue}0.9656&38.19 / 0.9518&{\color{blue}36.80} / 0.9609&{\color{blue}36.62} / 0.9590
&32.94 / 0.9309&{\color{red}33.09} / \color{blue}0.9380&35.59 / 0.9208&{\color{blue}34.15} / 0.9365&{\color{blue}33.94} / 0.9315\\
Ours-3D, $3\times3\times3$
&{\color{red}36.02} / {\color{red}0.9618}& {\color{blue}35.80} / \color{red}0.9666& {\color{red}38.78} / \color{red}0.9580& {\color{red}37.04} / \color{red}0.9624& {\color{red}36.91} / \color{red}0.9622
&{\color{red}33.29} / \color{red}0.9372& {\color{blue}33.05} / \color{red}0.9400& {\color{red}36.17} / \color{red}0.9301& {\color{red}34.40} / \color{red}0.9390& {\color{red}34.23} / \color{red}0.9366\\
\hline
KPN~\cite{KPN}, $5\times5$, $\sigma$ blind 
&35.44 / 0.9577&35.03 / 0.9605&38.03 / 0.9506&36.30 / 0.9586&36.20 / 0.9569
&32.73 / 0.9302&32.36 / 0.9312&35.39 / 0.9185&33.61 / 0.9309&33.52 / 0.9277\\
Ours-3D, $3\times3\times3$, $\sigma$ blind 
&{\color{blue}35.70} / 0.9590&35.47 / 0.9633&{\color{blue}38.35} / 0.9538&36.67 / 0.9615&36.55 / 0.9594
&{\color{blue}33.02} / 0.9327&32.79 / 0.9348&{\color{blue}35.78} / 0.9239&34.09 / 0.9361&33.92 / 0.9319 \\